\documentclass{article}
\usepackage{ijcai22}
\usepackage{multirow}
\usepackage{times}
\usepackage{soul}
\usepackage{url}
\usepackage[hidelinks]{hyperref}
\usepackage[utf8]{inputenc}
\usepackage{subcaption}
\usepackage{graphicx}
\usepackage{amsmath}
\usepackage{amsthm}
\usepackage{booktabs}
\usepackage{algorithm}
\usepackage{algpseudocode}
\usepackage{amsfonts}
\usepackage{calligra}

\title{On the Equivalence of the Weighted Tsetlin Machine and the Perceptron}

\author{
Jivitesh Sharma
\and
Ole-Christoffer Granmo\And
Lei Jiao
\affiliations
\emph{Center for Artificial Intelligence Research, University of Agder, Norway}
\emails
\{jivitesh.sharma, ole.granmo, lei.jiao\}@uia.no
}

\begin{document}

\maketitle

\begin{abstract}
    Tsetlin Machine (TM) has been gaining popularity as an inherently interpretable machine leaning method that is able to achieve promising performance with low computational complexity on a variety of applications. The interpretability and the low computational complexity of the TM are inherited from the Boolean expressions for representing various sub-patterns.  Although possessing favorable properties, TM has not been the go-to method for AI applications, mainly due to its conceptual and theoretical differences compared with perceptrons and neural networks, which are more widely known and well understood.  In this paper, we provide detailed insights for the operational concept of the TM, and try to bridge the gap in the theoretical understanding between the perceptron and the TM. More specifically, we study the operational concept of the TM following the analytical structure of perceptrons, showing the resemblance between the perceptrons and the TM. Through the analysis, we indicated that the TM's weight update can be considered as a special case of the gradient weight update. We also perform an empirical analysis of TM by showing the flexibility in determining the clause length, visualization of decision boundaries and obtaining interpretable boolean expressions from TM. In addition, we also discuss the advantages of TM in terms of its structure and its ability to solve more complex problems.
\end{abstract}

\section{Introduction}
Researchers across various fields are increasingly paying attention to the interpretability of AI techniques. While interpretability previously was inherent in most machine learning approaches, the state-of-the-art methods now increasingly rely on black-box deep neural networks (DNNs). Natively, DNNs can hardly be interpreted during the learning stage or while producing outputs~\cite{nnfragile_vvimp}. A surge of techniques attempts to open the black box by visual explanations and gradient-based interpretability~\cite{saliency,interpretcnns,patchnet,dissection,gradcam}, but do not change the black-box nature. \\
The Tsetlin Machine (TM) is a natively interpretable rule-based machine learning algorithm that produces logical rules~\cite{granmo2018}. Despite being logic-based, the TM is a universal function approximator, like a neural network. In brief, it employs an ensemble of Tsetlin Automata (TA) that learns propositional logic expressions from Boolean input features. Propositional logic drives learning, eliminating the requirement for floating-point operations. Due to its Boolean representations and finite-state automata learning mechanisms, it has a minimalistic memory footprint.  More importantly, TM achieves interpretability by leveraging sparse disjunctive normal form.  Indeed, humans are particularly good at understanding flat and short logical AND-rules, reflecting human reasoning~\cite{human-reasoning}.\\
Because the operational concept of TMs is significantly different from that of neural networks, TM is really challenging for those who are used to the neural networks to understand. For this reason, the TMs have not been considered as the to-go method in the machine learning society. In this paper, we show the operational concept of TMs for its learning phase, following the structure that is widely used in the analysis of the neural networks. Particularly, we aim at showing the resemblance between the two distinct techniques and reveal the concept and the advantages of the TMs in a painless manner. In more details, we divide TM's learning into two phases: the clause learning phase and the clause weight update phase.  \\
For the clause learning phase, we show that one clause can learn one or multiple sub-patterns given enough updates. This phase bears resemblance to the connections and the activation functions in a perceptron. For the clause weight update phase, we indicate that the clauses are weighted according to their correctness, which is similar to the weights in a perceptron. For this reason, following the concept of the perceptron convergence theorem, the clause weight update for the TM can also be confirmed. To summarize, the clauses can learn any sub-patterns from data and the clause weights show the importance of such sub-patterns, similar to the connections and the weights in a perceptron. In addition, we visualize the decision boundaries using clauses of TM and formalize its advantages over other the perceptron.

\section{Review of the Tsetlin Machine}
\label{sec:TM}
The Tsetlin Machine is a machine learning algorithm based on Boolean expressions called clauses that individually identify sub-patterns in data and are aggregated together as a weighted sum of Boolean inputs.\\
\begin{figure}[h]
  \begin{center}
  \includegraphics[width=\linewidth]{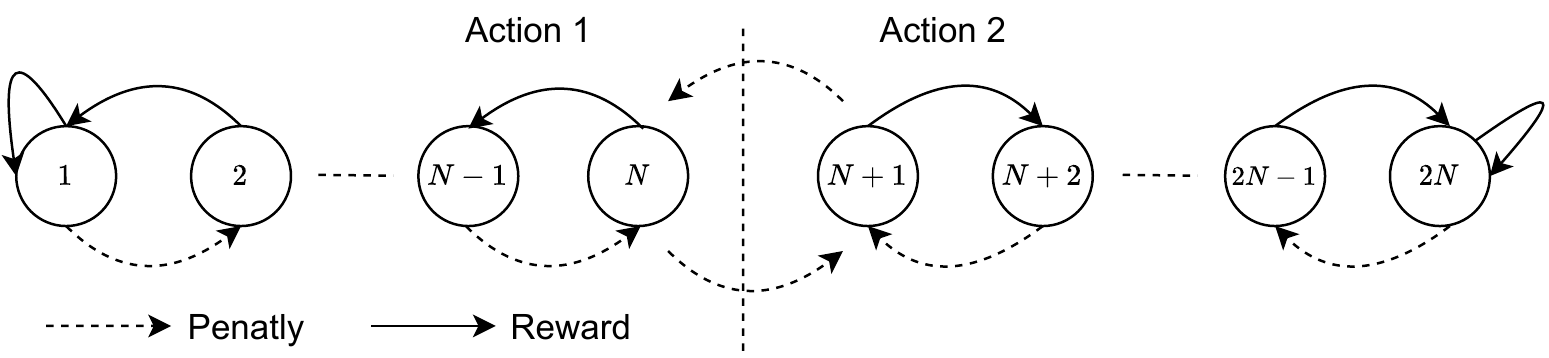}
  \caption{A two-action Tsetlin Automaton with $2N$ states.}\label{figTA}
  \end{center}
\end{figure}

\begin{figure}[ht]
\begin{center}
\centerline{\includegraphics[width=\columnwidth]{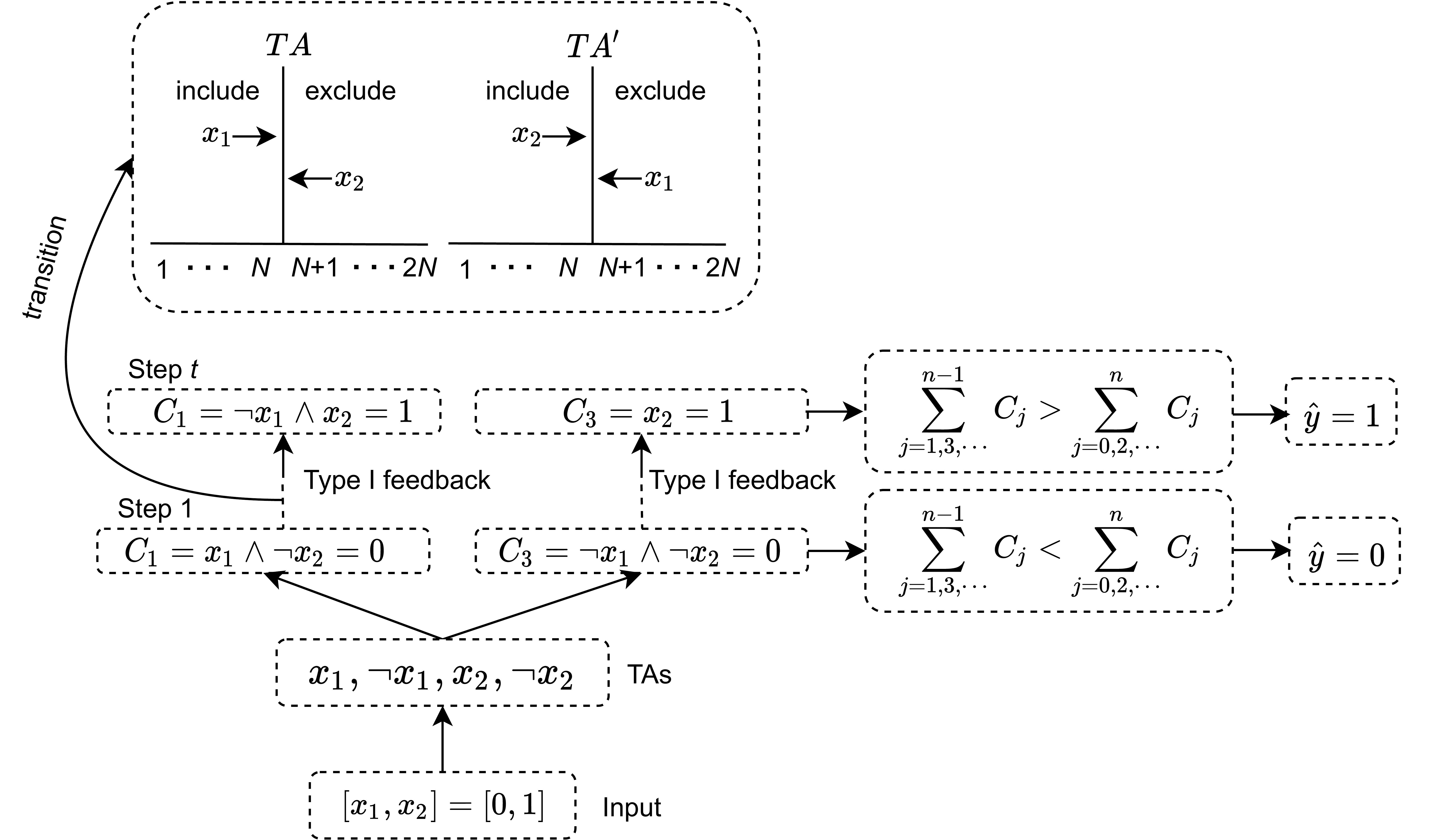}}
\caption{TM learning dynamics for an XOR-gate training sample, with input ($x_1=0, x_2=1$) and output target $y=1$.}
\label{figure:tm_architecture_basic}
\end{center}
\end{figure}
\paragraph{Structure.} A TM in its simplest form takes a feature vector $\mathbf{x} = [x_1, x_2, \ldots, x_o] \in \{0,1\}^o$ of $o$ propositional values as input and assigns the vector a class $\hat{y} \in \{0,1\}$.  In brief, the input vector $\mathbf{x}$ provides the literal set $L = \{l_1, l_2, \ldots, l_{2o}\} = \{x_1, x_2, \ldots, x_{o}, \lnot x_1, \lnot x_2, \ldots, \lnot x_o\}$, consisting of the input features and their negations. By selecting subsets $L_j \subseteq L$ of the literals, the TM can build arbitrarily complex patterns, ANDing the selected literals to form conjunctive clauses:
\begin{equation}
\label{eq:1}
C_j(\mathbf{x})= \bigwedge_{l_k \in L_j} l_k.
\end{equation}
Above, $j \in \{1, 2, \ldots, n\}$ is the index of a clause $C_j$ and $k \in \{1, 2, \ldots, 2o\}$ refers to a particular literal $l_k$. $n$ is the total number of clauses.  As an example, the clause $C_j(\mathbf{x}) = x_1 \land \lnot x_2$ consists of the literals $L_j = \{x_1, \lnot x_2\}$ and evaluates to $1$ when $x_1=1$ and $x_2=0$. \\
The TM assigns one Tsetlin Automata (TA) \cite{ta,LA_book} per literal $l_k$ per clause $C_j$ to build the clauses. The TA assigned to literal $l_k$ of clause $C_j$ decides whether $l_k$ is \emph{Excluded} or \emph{Included} in $C_j$. Figure~\ref{figTA} depicts a two-action TA with $2N$ states.  For states $1$ to $N$, the TA performs action \emph{Exclude} (Action 1), while for states $N + 1$ to $2N$ it performs action \emph{Include} (Action 2). As feedback to the action performed, the environment responds with either a Reward or a Penalty. If the TA receives a Reward, it moves deeper into the side of the action. If it receives a Penalty, it moves towards the middle and eventually switches action.\\
With $n$ clauses and $2o$ literals, we have in total $n\times2o$ TAs. We organize the states of these in an $n\times2o$ matrix $A = [a_k^j] \in \{1, 2, \ldots, 2N\}^{n\times2o}$. We will use the function $g(\cdot)$ to map the automaton state $a_k^j$ to Action $0$ (\emph{Exclude}) for states $1$ to $N$ and to Action $1$ (\emph{Include}) for states $N+1$ to~$2N$:
\begin{equation}
\label{eq:g}
g(a_k^j) = \begin{cases}
1& a_k^j > N\\
0& otherwise.
\end{cases}
\end{equation}
We then can connect the states $a_k^j$ of the TAs assigned to clause $C_j$ with its composition as follows:
\begin{equation}
\label{eq:2}
C_j(\mathbf{x}) = \bigwedge_{l_k \in L_j} l_k = \bigwedge_{k=1}^{2o} \left[g(a_k^j) \Rightarrow l_k\right].
\end{equation}
Here, $l_k$ is one of the literals and $a_k^j$ is the state of its TA in clause $C_j$. The logical \emph{imply} operator~$\Rightarrow$ implements the \emph{Exclude}/\emph{Include} action. That is, the 
\emph{imply} operator is always $1$ if $g(a_k^j)=0$ (\emph{Exclude}), while if $g(a_k^j)=1$ (\emph{Include}) the truth value is decided by the truth value of the literal.
\paragraph{Classification.} Classification is performed as a majority vote. The odd-numbered half of the clauses vote for class $\hat{y} = 0$ and the even-numbered half vote for $\hat{y} = 1$:
\begin{equation}
    \hat{y} = 0 \le \sum_{j=1,3,\ldots}^{n-1} \bigwedge_{k=1}^{2o} \left[g(a_k^j) \Rightarrow l_k\right] - \sum_{j=2,4,\ldots}^{n} \bigwedge_{k=1}^{2o} \left[g(a_k^j) \Rightarrow l_k\right]. \label{eqn:prediction}
\end{equation}
As such, the odd-numbered clauses have positive polarity, while the even-numbered ones have negative polarity. As an example, consider the input vector $\mathbf{x} = [0, 1]$ in the lower part of Figure \ref{figure:tm_architecture_basic}. The figure depicts two clauses of positive polarity, $C_1(\mathbf{x}) = x_1 \land \lnot x_2$ and $C_3(\mathbf{x}) = \lnot x_1 \land \lnot x_2$ (the negative polarity clauses are not shown). Both of the clauses evaluate to zero, leading to class prediction $\hat{y} = 0$.
\begin{table}[t]
\centering
\vskip 0.15in
\begin{center}
\begin{small}
\begin{tabular}{l|l|l|l}
    \hline
    \multirow{2}{*}{Input}&Clause & \ \ \ \ \ \ \ 1 & \ \ \ \ \ \ \ 0 \\
    &{Literal} &\ \ 1 \ \ \ \ \ \ 0 &\ \ 1 \ \ \ \ \ \ 0 \\
    \hline
    \multirow{2}{*}{Include Literal}&P(Reward)&$\frac{s-1}{s}$\ \ \ NA & \ \ 0 \ \ \ \ \ \ 0\\ [1mm]
    &P(Inaction)&$\ \ \frac{1}{s}$\ \ \ \ \ NA &$\frac{s-1}{s}$ \ $\frac{s-1}{s}$ \\ [1mm]
    &P(Penalty)& \ \ 0 \ \ \ \ \ NA& $\ \ \frac{1}{s}$ \ \ \ \ \  $\frac{1}{s}$ \\ [1mm]
    \hline
    \multirow{2}{*}{Exclude Literal}&P(Reward)& \ \ 0 \ \ \ \ \ \ $\frac{1}{s}$ & $\ \ \frac{1}{s}$ \ \ \ \ \
    $\frac{1}{s}$ \\ [1mm]
    &P(Inaction)&$ \ \ \frac{1}{s}$\ \ \ \ $\frac{s-1}{s}$  &$\frac{s-1}{s}$ \ $\frac{s-1}{s}$ \\ [1mm]
    &P(Penalty)&$\frac{s-1}{s}$ \ \ \ \ 0& \ \ 0 \ \ \ \ \ \ 0 \\ [1mm]
    \hline
\end{tabular}
\end{small}
\end{center}
\caption{Type-I Feedback}
\label{table:type_i}
\end{table}

\begin{table}[t]
\centering
\vskip 0.15in
\begin{center}
\begin{small}
\begin{tabular}{l|l|l|l}
    \hline
    \multirow{2}{*}{Input}&Clause & \ \ \ \ \ \ \ 1 & \ \ \ \ \ \ \ 0 \\
    &{Literal} &\ \ 1 \ \ \ \ \ \ 0 &\ \ 1 \ \ \ \ \ \ 0 \\
    \hline
    \multirow{2}{*}{Include Literal}&P(Reward)&\ \ 0 \ \ \ NA & \ \ 0 \ \ \ \ \ \ 0\\[1mm]
    &P(Inaction)&1.0 \ \  NA &  1.0 \ \ \ 1.0 \\[1mm]
    &P(Penalty)&\ \ 0 \ \ \ NA & \ \ 0 \ \ \ \ \ \ 0\\[1mm]
    \hline
    \multirow{2}{*}{Exclude Literal}&P(Reward)&\ \ 0 \ \ \ \ 0 & \ \ 0 \ \ \ \ \ \ 0\\[1mm]
    &P(Inaction)&1.0 \ \ \ 0 &  1.0 \ \ \ 1.0 \\[1mm]
    &P(Penalty)&\ \ 0 \ \  1.0 & \ \ 0 \ \ \ \ \ \ 0\\[1mm]
    \hline
\end{tabular}
\end{small}
\end{center}
\caption{Type-II Feedback}
\label{table:type_ii}
\end{table}
\paragraph{Learning.} The upper part of Figure \ref{figure:tm_architecture_basic} illustrates learning. A TM learns online, processing one training example $(\mathbf{x}, y)$ at a time. Based on $(\mathbf{x}, y)$, the TM rewards or penalizes its TAs, which amounts to increasing or decreasing their states. There are two kinds of feedback: Type I Feedback produces frequent patterns and Type II Feedback increases the discrimination power of the patterns.\\
Type I feedback is given stochastically to clauses with positive polarity when $y=1$  and to clauses with negative polarity when $y=0$. Conversely, Type II Feedback is given stochastically to clauses with positive polarity when $y=0$ and to clauses with negative polarity when $y=1$. The probability of a clause being updated is based on the vote sum $v$: $v = \sum_{j=1,3,\ldots}^{n-1} \bigwedge_{k=1}^{2o} \left[g(a_k^j) \Rightarrow l_k\right] - \sum_{j=2,4,\ldots}^{n} \bigwedge_{k=1}^{2o} \left[g(a_k^j) \Rightarrow l_k\right]$. The voting error is calculated as:
\begin{equation}
\label{eq:TT}
\epsilon = \begin{cases}
T-v,& for~ y=1\\
T+v,& for~ y=0.
\end{cases}
\end{equation}
Here, $T$ is a user-configurable voting margin yielding an ensemble effect. The probability of updating each clause is $P(\mathrm{Feedback}) = \frac{\epsilon}{2T}$.  Random sampling from $P(\mathrm{Feedback})$ will decided which clauses to update, and then the following TA state updates can be formulated as matrix additions, subdividing Type I Feedback into feedback Type Ia and Type Ib:
\begin{equation}
    A^*_{t+1} = A_t + F^{\mathit{II}} + F^{Ia} - F^{Ib}.
    \label{eqn:learning_step_1}
\end{equation}
Here, $A_t = [a^j_k] \in \{1, 2, \ldots, 2N\}^{n \times 2o}$ contains the states of the TAs at time step $t$ and $A^*_{t+1}$ contains the updated state for time step $t+1$ (before clipping). The matrices $F^{\mathit{Ia}} \in \{0,1\}^{n \times 2o}$ and $F^{\mathit{Ib}} \in \{0,1\}^{n \times 2o}$ contain Type I Feedback. A zero-element means no feedback and a one-element means feedback. As shown in Table \ref{table:type_ii} on the left, two rules govern Type I feedback:
\begin{itemize}
    \item \textbf{Type Ia Feedback} is given with probability $\frac{s-1}{s}$ whenever both clause and literal are $1$-valued\footnote{Note that the probability $\frac{s-1}{s}$ is replaced by $1$ when boosting true positives.}. It penalizes \emph{Exclude} actions and rewards \emph{Include} actions. The purpose is to remember and refine the patterns manifested in the current input $\mathbf{x}$. This is achieved by moving the state of the TA toward the right side. The user-configurable parameter $s$ controls pattern frequency, i.e., a higher $s$ produces less frequent patterns.
    \item \textbf{Type Ib Feedback} is given with probability $\frac{1}{s}$ whenever either clause or literal is $0$-valued. This feedback rewards \emph{Exclude} actions and penalizes \emph{Include} actions to coarsen patterns, combating overfitting. Thus, the selected TA states are decreased.
\end{itemize}
The matrix $F^{\mathit{II}} \in \{0, 1\}^{n \times 2o}$ contains Type II Feedback to the TAs, given per Table \ref{table:type_ii} on the right.
\begin{itemize}
\item \textbf{Type II Feedback} penalizes \emph{Exclude} actions to make the clauses more discriminative, combating false positives. That is, if the literal is $0$-valued and the clause is $1$-valued, TA that has the current state below $N+1$ are encouraged to move towards right side. Eventually the clause becomes $0$-valued for that particular input, upon inclusion of the $0$-valued literal.
\end{itemize}
The final updating step for training example  $(\mathbf{x}, y)$ is to clip the state values to make sure that they stay within value $1$ and $2N$:
\begin{equation}
    A_{t+1} = \mathit{clip}\left(A^*_{t+1}, 1, 2N\right). \label{eqn:learning_step_2}
\end{equation}

For example, both of the clauses in Figure \ref{figure:tm_architecture_basic} receives Type~I Feedback over several training examples, making them resemble the input associated with $y=1$.

\subsection{Weighted Tsetlin Machine}
In this subsection, we detail the TM with weights. The learning of weights is based on increasing the weight of clauses that receive Type Ia feedback (due to true positive output) and decreasing the weight of clauses that receive Type II feedback (due to false positive output). The overall rationale is to determine which clauses are inaccurate and thus must team up to obtain high accuracy as a team (low weight clauses), and which clauses are sufficiently accurate to operate more independently (high weight clauses). The weight updating procedure is summarized in Algorithm \ref{algo:tm}. Here, $w_i$ is the weight of clause $C_i$ at the $n^{th}$ training round (ignoring polarity to simplify notation). The first step of a training round is to calculate the clause output as per Equation~(\ref{eq:2}). The weight of a clause is only updated if the clause output $C_i$ is 1 and the clause has been selected for feedback ($P_i$ = 1). Then the polarity of the clause and the class label $y$ decide the type of feedback given. That is, like a regular TM, positive polarity clauses receive Type Ia feedback if the clause output is a true positive, and similarly, they receive Type II feedback if the clause output is a false positive. For clauses with negative polarity, the feedback types switch roles. When clauses receive Type Ia or Type II feedback, their weights are updated accordingly. We use the stochastic searching on the line (SSL) automaton to learn appropriate weights. SSL is an optimization scheme for unknown stochastic environments~\cite{oommen}. The goal is to find an unknown location $\lambda^*$ within a search interval $[0,1]$. In order to find $\lambda^*$, the only available information for the Learning Mechanism (LM) is the possibly faulty feedback from its attached environment $E$.\\
In SSL, the search space $\lambda$ is discretized into $N$ points, $\{0,1/N,2/N,...,(N-1)/N,1\}$ with N being the discretization resolution. During the search, the LM has a location $\lambda \in \{0,1/N,2/N,...,(N-1)/N,1\}$, and can freely move to the left or to the right from its current location. The environment $E$ provides two types of feedback: $E = 1$ is the environment suggestion to increase the value of $\lambda$ by one step, and $E = 0$ is the environment suggestion to decrease the value of $\lambda$ by one step. The next location of $\lambda$, i.e. $\lambda_{n + 1}$, can thus be expressed as follows:
\begin{equation}
 \lambda_{n + 1} =
    \begin{cases}
      \lambda_n+1/N, & \text{if $E_n=1$,}\\
      \lambda_n-1/N, & \text{if $E_n=0$.}\\
    \end{cases}       
\end{equation}
\begin{equation}
 \lambda_{n + 1} =
    \begin{cases}
      \lambda_n, & \text{if $\lambda_n=1$ and $E_n=1$,}\\
      \lambda_n, & \text{if $\lambda_n=0$ and $E_n=0$.}\\
    \end{cases}       
\end{equation}
Asymptotically, the learning mechanics is able to find a value arbitrarily close to $\lambda^*$ when $N\rightarrow\infty$ and $n\rightarrow\infty$. In our case, the search space of clause weights is $[0, \infty]$, so we use resolution $N = 1$, with no upper bound for $\lambda$. Accordingly, we operate with integer weights. As in Algorithm~\ref{algo:tm}, if the clause output is a true positive, we simply increase the weight by $1$. Conversely, if the clause output is a false positive, we decrease the weight by $1$.\\
By following the above procedure, the goal is to make low precision clauses team up
by giving them low weights, so that they together can reach the summation target $T$. By teaming up, precision increases due to the resulting ensemble effect. Clauses with high precision, however, obtain a higher weight, allowing them to operate more independently.\\
The above weighting scheme has several advantages. First of all, increment and decrement operations on integers are computationally less costly than multiplication based updates of real-valued weights. Additionally, a clause with an integer weight can be seen as multiple copies of the same clause, making it more interpretable than real-valued weighting, as shown in the next section. Additionally, clauses can be turned completely off by setting their weights to $0$ if they do not contribute positively to the classification task. For a more detailed explanation of the weighted TM, please refer to ~\cite{abeyrathna2021integer}.

\begin{algorithm}[t]
\caption{Complete WTM learning process}\label{algo:tm}
\begin{algorithmic}[1]
\State \textbf{Input:} Training data batch $(B, x, y) \quad \rhd B \geq 1$
\State \textbf{Initialize:} Random initialization of TAs
\State \textbf{Begin:} $n^{th}$ training round
\For{$i = 1, ...,m$} \textbf{if} $P_i = 1$  
    \If{($y = 1$ \textbf{and} $i$ is odd) \textbf{or} ($y = 0$ \textbf{and} $i$ is even)}
        \If{$c_i = 1$} 
            \State $w_i \leftarrow w_i+1$ 
            \For{feature $k=1,...,2o$}
                \If{$l_k = 1$}
                    \State Type Ia Feedback
                \Else:
                    \State Type Ib Feedback
                \EndIf
            \EndFor
        \Else:
            \State $w_i \leftarrow w_i \quad \rhd$ \text{[No Change]}
            \State Type Ib Feedback
        \EndIf
    \Else: ($y = 1$ \textbf{and} $i$ is even) \textbf{or} ($y = 0$ \textbf{and} $i$ is odd)
        \If{$c_i = 1$} 
            \If{$w_i > 0$}
                \State $w_i \leftarrow w_i-1$ 
            \EndIf
            \For{feature $k=1,...,2o$}
                \If{$l_k = 0$}
                    \State Type II Feedback
                \Else:
                    \State Inaction
                \EndIf
            \EndFor
        \Else:
            \State $w_i \leftarrow w_i \quad \rhd$ \text{[No Change]}
            \State Inaction
        \EndIf
    \EndIf
\EndFor
\end{algorithmic}
\end{algorithm}

\section{Convergence Analysis of the Tsetlin Machine}
\label{sec:converge}
\begin{figure}[b]
    \centering
    \includegraphics[width=0.5\textwidth]{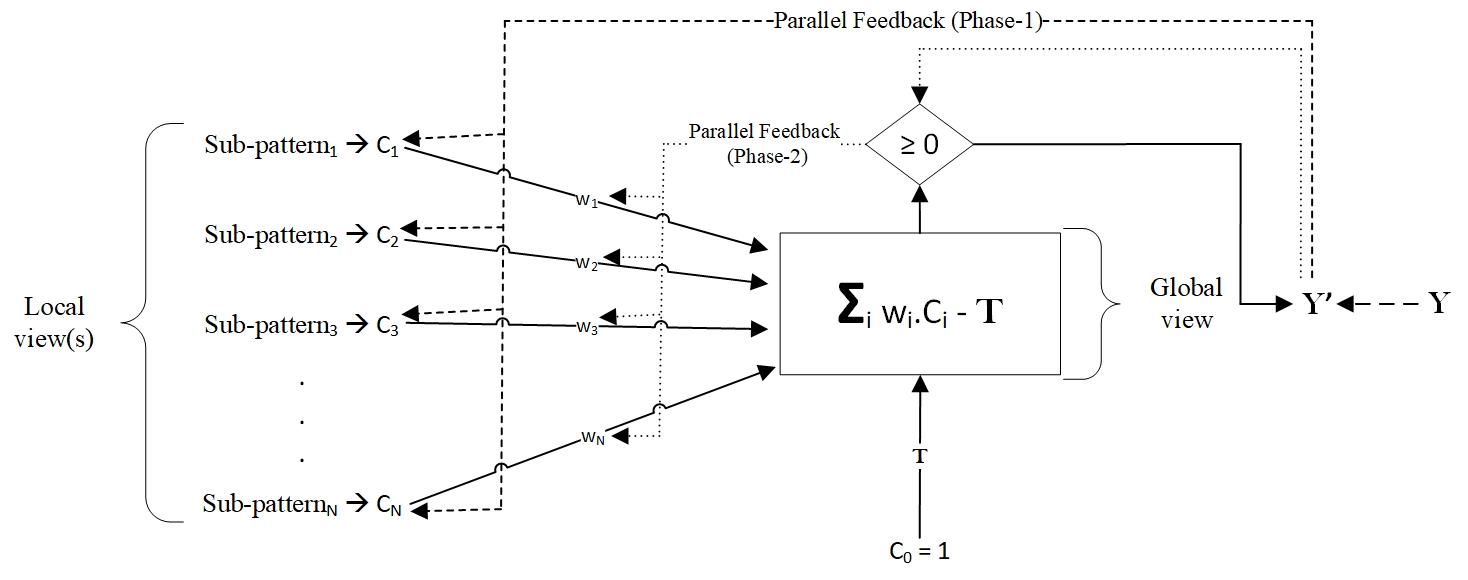}
    \caption{Phase-1 and Phase-2 Feedbacks}
    \label{fig:tm}
\end{figure}
As discussed in Section \ref{sec:TM}, TM learns clauses to identify sub-patterns in data. These sub-patterns are aggregated using a linearly weighted sum of clauses, where the weights depend on how well the clauses detect sub-patterns. Now, let's consider the weighted TM with clause weights $w_i$ and clauses $C_i$, where even $i$ represent negative polarity clauses and odd $i$ represent positive polarity clauses:
\begin{equation}
    \sum_i w_iC_i \, \geq \, T \; \rightarrow \;  \hat{y} \tag{Using Eq.~(\ref{eq:TT})}
\end{equation}
\begin{equation}
    \sum_i w_i\left(\bigwedge_{l_k \in L_i} l_k \right) \, \geq \, T \; \rightarrow \;  \hat{y} \tag{Using Eq.~(\ref{eq:1})}
\end{equation}
The learning of clauses and its weights can be considered as separate phases as described in Section \ref{sec:TM} and shown in Algorithm \ref{algo:tm}. Each $w_i$ is updated according to its associated clause $C_i$'s correctness. Whereas each clause $C_i$ obtains feedback by comparing the clause output (and its polarity) and the true label. So, the convergence of TM is divided into two phases:
\begin{itemize}
    \item \textbf{Phase $\mathbf{1}$: Clause learning} - \emph{Local sub-pattern learning}. A single clause is able to learn a single correct sub-pattern based on the feedback given to it by comparing the clause output with the desired output. This is shown according to parallel feedback given to the TAs. (Local view, see Subsection \ref{sec:cl})
    \item \textbf{Phase $\mathbf{2}$: Clause weight update} - \emph{Global pattern aggregation}. The sub-patterns learnt by the clauses are weighted according to their correctness, i.e. a weighted sum of sub-patterns. The weighted TM and its weight update is shown to be akin to the perceptron. (Global view, see Subsection \ref{sec:w})
\end{itemize}

\subsection{Clause learning}
\label{sec:cl}
In this subsection, we illustrate that one clause can capture one individual sub-pattern or several sub-patterns if the sub-patterns can be represented by the clause jointly.

To clarify the meaning of capturing one individual sub-pattern or several sub-patterns jointly, we look at the XOR and the AND operators as examples.  
Let us review the case where one clause captures an individual sub-pattern by looking at the XOR operator. For this operator, ($x_1=1$, $x_2=0$) and ($x_1=0$, $x_2=1$) give $y=1$  while $y=0$ otherwise. Clearly, we have two sub-patterns and the input bits of the sub-patterns are mutual exclusive. For this reason, we need one clause $C_1=x_1\wedge \neg x_2$ to capture the first sub-pattern and $C_2=\neg x_1\wedge x_2$ to represent the second one. Obviously, there is no possibility to represent the two sub-patterns jointly by one clause, and thus one clause must correspond to each sub-pattern. Indeed, the TM can learn almost surely the intended logic in infinite time horizon. The convergence of the XOR operator has been proven in~\cite{jiao2021convergence}. 

For the case where sub-patterns can be presented jointly, we exam the AND operator. Clearly, in addition to ($x_1=1$, $x_2=0$) and ($x_1=0$, $x_2=1$), ($x_1=1$, $x_2=1$) will also trigger a positive output. Different from the XOR case where the sub-patterns are mutually exclusive, the latter two sub-patterns in the AND operator can be jointly represented by $C_1=x_2$. Although in the AND operator three sub-patterns exist, two clauses, e.g., $C_1=x_2$ and $C_2=x1\wedge\neg x_2$, are sufficient to present the intended AND operator.  Indeed, the TM can learn almost surely the intended AND operator in infinite time horizon. The convergence of the operator has been proven in~\cite{jiao2021convergenceAND}. 

From the above mentioned two examples, we can see that a clause can indeed learn either a sub-pattern individually or multiple sub-patterns jointly for those special cases where the input is 2-bits long. In general, we conjuncture that a clause, after learning, can capture one or multiple sub-patterns from the training sample when the input has more bits. The proof of the general case is not trivial because the feedback for a certain literal is not only determined by its own state, but also by the output of the clause that is jointly determined by all its literals.  Although this conjuncture has not been theoretically proven, we have observed from simulations that the clauses in a TM can indeed present sub-patterns efficiently. In what follows, we present the clause learning phase formally, aiming at revealing the dynamics of the learning and providing more insights for a better comprehension.

TM learns to recognize local patterns courtesy of clauses, which are propositional expressions of binarized features (literals), in original or negated form, connected by logical \emph{AND} operations. Correct sub-patterns are learnt by updating TA states that are associated with each literal which thereafter results in updated clauses. Consider a sub-pattern or a group of sub-patterns that one clause can learn joint. For the weighted TM, output is expected to be greater than $T$ once learnt for the intended sub-pattern (or joint sub-patterns), i.e.,  
\begin{equation}
    \label{eq:utm}
    w_i \bigwedge_{k=1}^{2o} \left[g(a_k^i) \Rightarrow l_k\right] \, \geq \, T \; \rightarrow \; \hat{y}
\end{equation}
Each TA $g(a_k^i)$, associated with a particular literal $l_k$, is updated according to the feedback given to the clause. The update of $w_1$ is to be discussed in the next Subsection. For updating the clause itself, we have  
\begin{equation}
    \label{eq:cf}
    C_i=\bigwedge_{k=1}^{2o} \left[g(a_k^i) \Rightarrow l_k\right] \, \xleftarrow[]{\text{Feedback}} \, \mathcal{F}(y == \hat{y})
\end{equation}
where $\mathcal{F}$ is the feedback given to the clause by comparing $\hat{y}$, the clause output and $y$, the true label. This feedback is sent to each literal. Note that once the feedback is given, the state updating process for each TA is independent and thus the TAs can be updated in parallel.

As the literals within the clause have an \emph{AND} relationship, any literal that produces 0 will results in a 0 for the clause output. On the contrary, the clause output 1 only when all literals output 1. This nature will result in a 0 value for a clause most probably when we randomly initiate the states of the TAs in the beginning of the learning. The 0 literal value or clause value will result in a Type Ib feedback for any ``true" training samples ($y=1$), which encourages the literals to be excluded. As the literals become excluded, the length of the clause is reduced. Once the 0-valued literals are all excluded and only the 1-valued literals are left, Type Ia feedback will come to the play and thus encourage more literals to be included in the clause. At the same time, Type~II feedback will depress possible false positive by including necessary literals upon a false training sample ($y=0$). This process will go back and forth during the learning process, until $w_iC_i$ reaches $T$. Once $w_iC_i\geq T$ holds,  the input of the sub-pattern is blocked by the TM, as per Eq.~(\ref{eq:TT}).      

\subsection{Clause weight update}
\label{sec:w}
As the clause is update to learn the intended sub-pattern, each clause is weighted according its importance and correctness. Here, the importance of the sub-pattern depends on its frequency of occurrence in different instances of data and its distinguishing capabilities for a particular task. For example, a clause that captures the sub-pattern of facial features in a ``dog vs cat'' task will have higher weights than a clause that captures sub-patterns of the tail, since facial feature are more discriminative for this task. The weighted TM can be represented as follows:
\begin{equation*}
    \sum_i w_iC_i \geq T,
\end{equation*}
which can be rewritten as:
\begin{equation}
    \label{eq:0}
    \sum_i w_iC_i -T \geq 0,
\end{equation}
where $w_i$ is the clause weight of clause $C_i$ and $T$ is the threshold. Clauses are represented in Equation~(\ref{eq:2}). This means that if this inequality holds for a particular input then it is assigned to class $1$, otherwise class $0$. For the case of TM, the threshold is static and assigned at the beginning of training. Here, we make $T$ learnable by reconstructing it as a clause weight. The clause associated with $T$ is a dummy clause whose output is always $1$. And, the weight $T$ is always subtracted from the weighted clause vote count. After those modifications, Eequation~(\ref{eq:0}) becomes analogous to a perceptron \cite{perceptron}, where $w_i$ are the weights, $C_i$ are the inputs and $T$ is the bias. Now, the weights of such a network are updated by the gradients of the error with respect to the weights. Let the error be represented as:
\begin{equation}
\label{eq:E}
    E = \left(\sum_i w_i.C_i -T\right) - y,     
\end{equation}
where, $y$ is the correct target. Clearly, the gradient of $E$ with respect to weight $w_i$ becomes:
\begin{equation}
    \nabla_{w_i} E = C_i,
\end{equation}
which means that the weight $w_i$ is updated as:
\begin{align}
\label{eq:4}
    w_i &= w_i + \eta\gamma\nabla_{w_i} E\\\nonumber
    &= w_i + \eta\gamma C_i,
\end{align}
where $\gamma \in \{-1,1\}$ represents correct prediction with $1$ and incorrect prediction with $-1$. This gives the direction of update and $\eta$ is the learning rate. In case of perceptron, $\eta=1$, which makes perceptron update:
\begin{align}
    w_i = w_i + C_i \quad \Rightarrow \quad  w_i = w_i + 1,       \tag{[since $C_i \in \{0,1\}$]}\\
    w_i = w_i - C_i \quad  \Rightarrow \quad  w_i = w_i - 1.   
    \label{eq:tmu}
\end{align}

This is how the clause weights are updated in weighted TM as shown in Section~\ref{sec:TM} and Algorithm~\ref{algo:tm}. From Algorithm~\ref{algo:tm}, we can see that, when the clause is correct, Equation~(\ref{eq:tmu}) is used to update the clause weight as shown in Lines $5-7$ and in case of the clause being incorrect, Equation~(\ref{eq:tmu}) is used to update the clause weight as shown in Lines $19-22$. The clause weights with clause output $C=0$ are not updated. \\

Algorithm~\ref{algo:tm} runs over all training instances (in batches or online manner) and the clause weights are incremented every time the clause detects the correct sub-pattern and decremented if it does not. As mentioned previously, the weights of clauses that capture important and more discriminative sub-patterns, i.e. ones that reduce the error $E$ and produce correct outputs, are incremented gradually and vice versa. 

Similarly, gradient of error in Equation~(\ref{eq:E}) with respect to $T$ is:
\begin{equation}
    \nabla_T E = - 1.
\end{equation}
Applying Equation~(\ref{eq:4}) to update $T$, we have:
\begin{equation}
\label{eq:8}
    T = T + \gamma(-1).
\end{equation}

From Equation~(\ref{eq:8}), we can see that $T$ is updated with the opposite polarity as $w_i$, which is what we expected as $T$ is always subtracted from the total vote count. This shows that the weight updating mechanism in the TM can be considered a special case of a perceptron with binary inputs (assuming the outputs of clauses are inputs here). Hence, the convergence properties and mathematical analysis of the perceptron~\cite{perceptron} also hold for the clause weight update of the TM, assuming the clauses as binary inputs. In addition, the clause weight update is equivalent to a gradient update with $\eta=1$. The convergence of perceptron is given as a reference point in Appendix~\ref{sec:app}.

\section{Empirical Analysis}
In this section, we present the empirical results based on our experiments. We firstly show the results for the clause learning phase with the focus on the relationship between the clause length and hyperparameter $s$. Thereafter, we consider the weights of the TM and show the resemblance between the neural networks and the TM.  
\subsection{Clause Length}
A clause evaluating to $1$ means that the sub-pattern associated with the clause is presented in a particular data instance and a clause evaluating to $0$ means the absence of the sub-pattern (but might to evaluate to $1$ for other instances, indicating presence of the sub-pattern). Understandably, if the length of a clause is longer, it learns more fine features or details in the sub-pattern. This is due to more literals are included for a longer clause so that the details/fine features can also be represented, with a cost of the risk of overfitting.  On the contrary, the short clause learns more generalized features and has easier readability. The length of the clause is depend on the nature of the problem, and also in part, depends on the $s$-parameter. The $s$-parameter is responsible for assigning the probabilities of reward, penalty and inaction for including and excluding literals (as given in Table~\ref{table:type_i}). From Table~\ref{table:type_i}, we can see that the probability on inclusion of a literal is high when $s$ is high, mainly due to Type Ia feedback.  To validate this statement, we show from Table~\ref{tab:s} the variation in clause size as a function of $s$-parameter. These clauses were obtained by training a TM on the Iris dataset for $50$ epochs. Clearly, longer clauses are found for a lager value of $s$. \\ 
\begin{table*}[h]
    \centering
    \begin{sc}
    \begin{tabular}{c|c|c}
    \hline
         & $s=10$  & $s=2$  \\
         \hline
\small{Clause-1:} & $\neg x_5 \land \neg x_{10} \land \neg x_{11} $   & $ \neg x_7 \land \neg x_8 \land \neg x_{10} $  \\
\small{Clause-2:} &  $ x_2 \land \neg x_0 \land \neg x_1 \land \neg x_4 \land \neg x_8 \land \neg x_9 \land \neg x_{10} \land \neg x_{15}$ & $ x_4 \land  x_7 \land \neg x_{11} $  \\
\small{Clause-3:} &  $ x_7 \land x_8 \land \neg x_{11} \land \neg x_{12} \land x_{15}$  & $ x_9 $               \\
\small{Clause-4:} &  $ x_1 \land \neg x_2 \land \neg x_6 \land x_7 $ & $ \neg x_0 \land \neg x_6 $         \\
\small{Clause-5:} &  $ x_7 \land \neg x_9 \land \neg x_{15} $ &  $x_{13} \land \neg x_0 \land \neg x_{14}$  \\
\small{Clause-6:} &  $ x_{10} \land \neg x_{11} \land \neg x_{13} \land x_{15}$   &    $ x_3 \land x_{12}  $       \\
\small{Clause-7:} &  $ \neg x_0 \land x_7 \land \neg x_{13} \land x_{15} \land \neg x_2$ & $\neg x_{10} $              \\
\small{Clause-8:} &  $ x_2 \land \neg x_0 \land \neg x_{13} $ & $ x_7 \land  x_9 \land  x_{10}  $ \\
\small{Clause-9:}&  $ x_7 \land \neg x_0 \land \neg x_6 \land \neg x_{13} $ & $ x_1 \land  x_{15} \land \neg x_{11} $ \\
\small{Clause-10:}&  $  x_{10} \land  x_{11} \land \neg x_0 \land \neg x_4 \land \neg x_{12} \land \neg x_{14} $ & $ x_{15} \land \neg x_7 $        \\
\small{Clause-11:}&  $ \neg x_5 \land \neg x_8 \land \neg x_9 \land \neg x_{12} $ & $ \neg x_1 \land \neg x_{10}   $      \\
\small{Clause-12:}&  $ \neg x_{15} $ &  $ x_0 \land \neg x_{11}   $     \\
\hline
    \end{tabular}
    \caption{Clause Size Variation with $s$-parameter.}
    \label{tab:s}
    \end{sc}
\end{table*}
A comparison between two different values of the $s$-parameter on memory consumption, training time and number of epochs required to reach $95\%$ accuracy on the Iris dataset is presented in Table \ref{tab:eval}. For this task, a TM consisting of $50$ clauses is employed. Clearly,  the TM with smaller $s$ value requires less memory due to the less included literals. For the same reason, a shorter training time per epoch is also achieved. Nevertheless, the smaller $s$ requires more training epochs to obtain the same accuracy, which requires slightly more overall training time. Even though these are different configurations of TM, both are capable of achieving comparable performance. The $s$-parameter is an important hyperparameter which needs to be carefully tuned to obtain optimal performance and interpretability. A detailed analysis and comparison of TM's memory and time consumption with other algorithms has been shown in~\cite{newcastle} and a more rigorous theoretical analysis of the $s$ parameter can be found in~\cite{onebit}.\\

\begin{table}[h]
    \centering
    \begin{sc}
    \begin{small}
    \begin{tabular}{c|c|c}
    \hline
          &   $s=10$  &   $s=2$  \\
          \hline
     \small{Memory (in Kb)}  &  204.8   &  122.5   \\
        \small{Epochs} & 250 &  350  \\
        \small{Training time per epoch (in millisec)} &    7.26  &  5.53    \\
    \hline
    \end{tabular}
    \end{small}
    \end{sc}
    \caption{Training Epochs and Memory Consumption to obtain $95\%$ accuracy on Iris.}
    \label{tab:eval}
\end{table}


\subsection{Visualization of Decision Boundaries}
\label{sec:vis}
\begin{figure*}[!h]
\begin{center}
\begin{subfigure}{0.3\textwidth}
  \centering
  \includegraphics[width=\columnwidth]{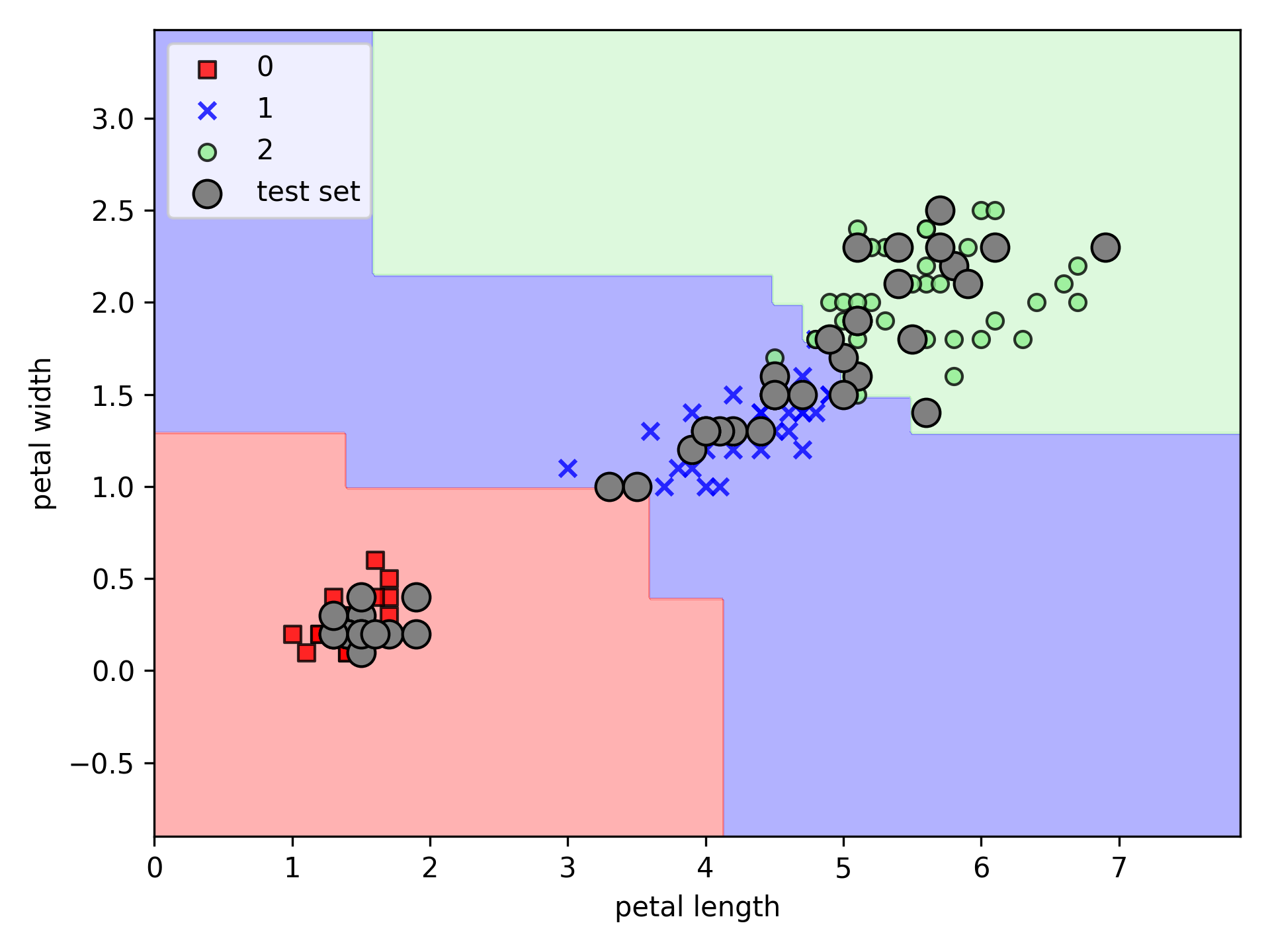}
  \caption{Tsetlin Machine.}
  \label{fig:sfig1}
\end{subfigure}
\begin{subfigure}{0.3\textwidth}
  \centering
  \includegraphics[width=\columnwidth]{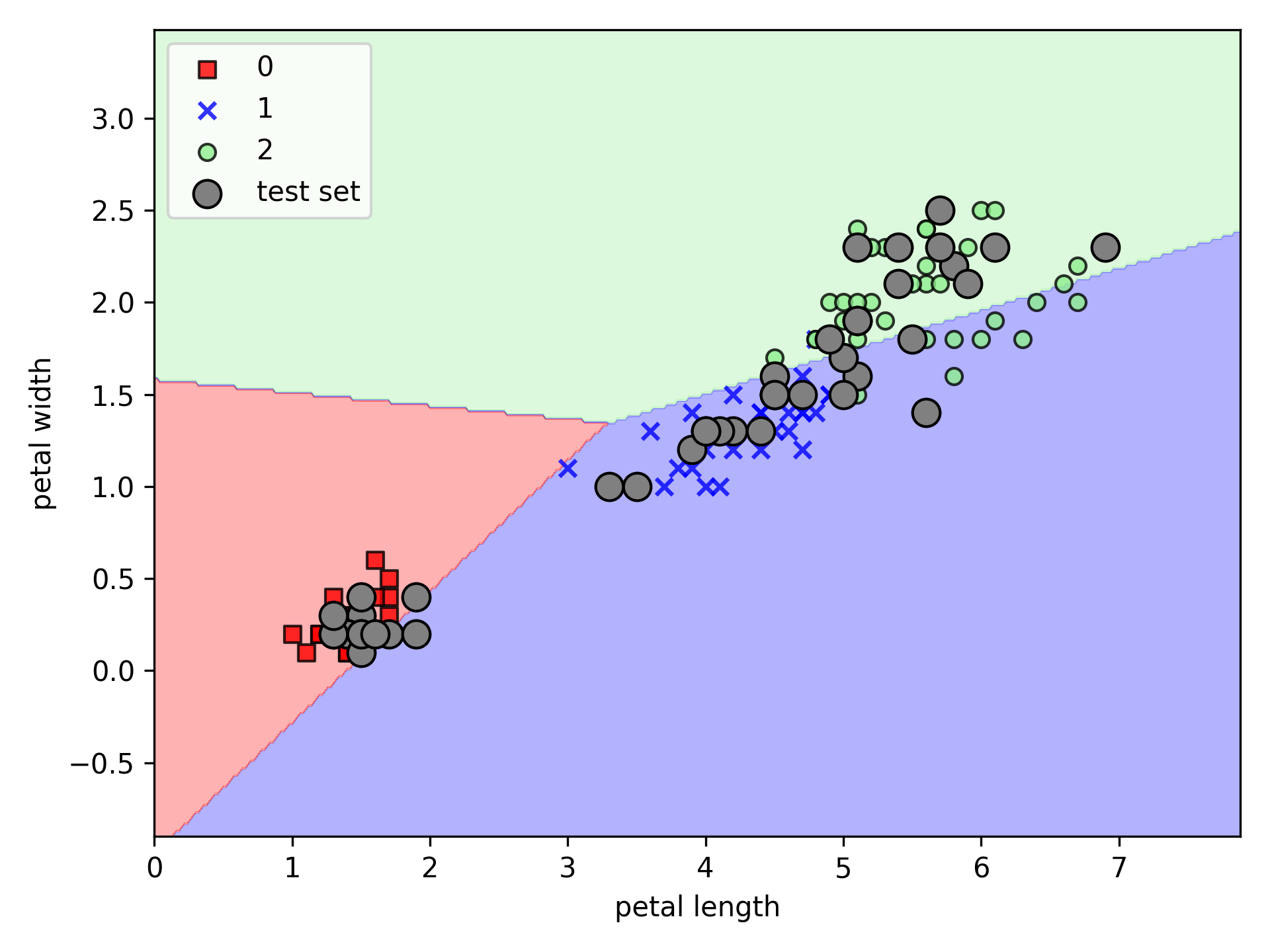}
  \caption{Perceptron.}
  \label{fig:sfig2}
\end{subfigure}
\begin{subfigure}{0.3\textwidth}
  \centering
  \includegraphics[width=\columnwidth]{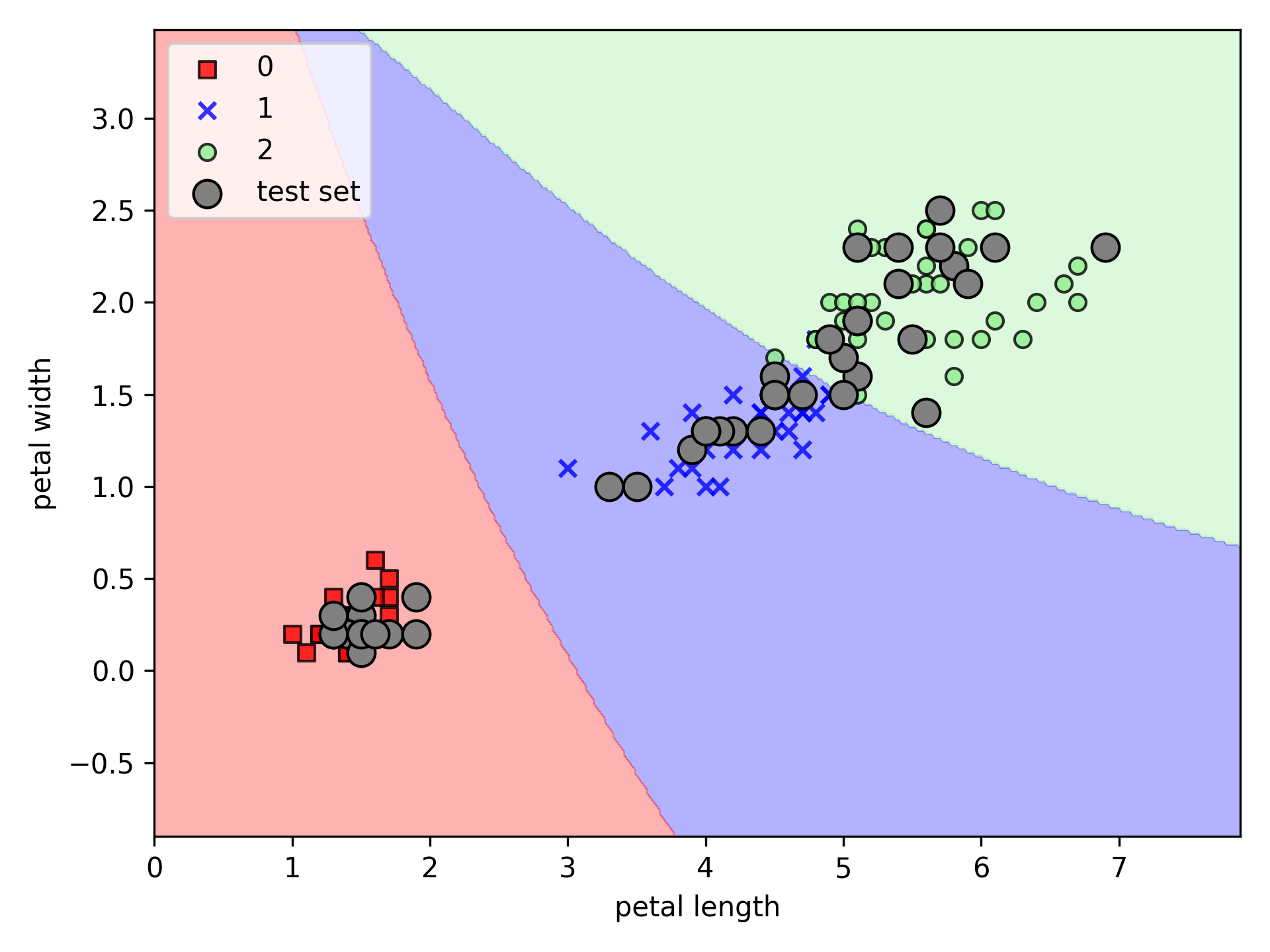}
  \caption{Single Layer NN.}
  \label{fig:sfig3}
\end{subfigure}%
\caption{Decision Boundary Visualization.}
\label{fig:vis}
\end{center}
\end{figure*}

\begin{figure}[!h]
    \centering
    \includegraphics[width=0.33\textwidth]{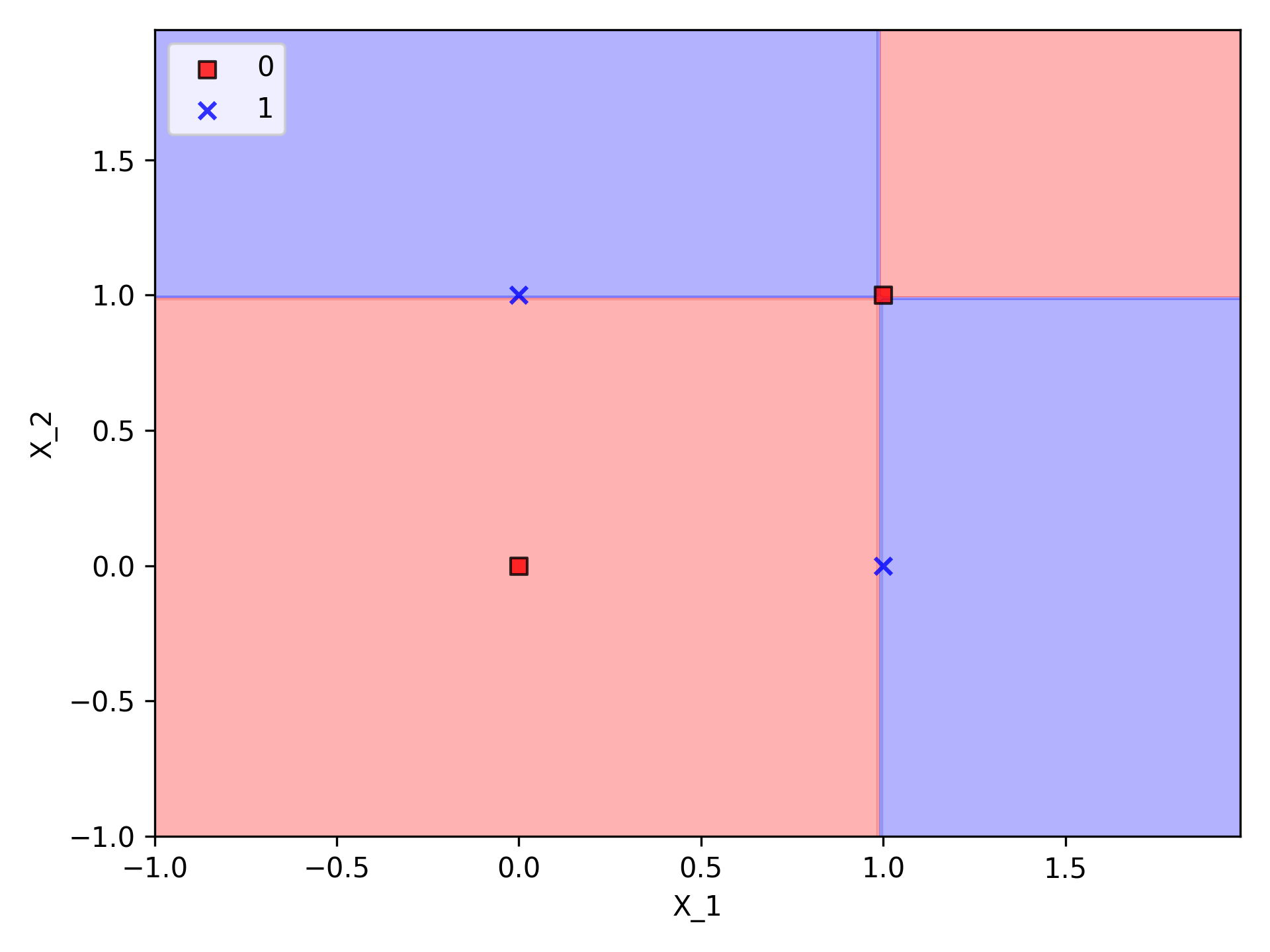}
    \caption{TM Decision Boundaries for the XOR problem.}
    \label{fig:xor_boundary}
\end{figure}

Visualizing decision boundaries is the one of the simplest ways of determining the pattern recognition abilities of an algorithm. It offers perspective about the decision function learnt by the method in order to best distinguish between patterns belonging to different categories. In this subsection, we visualize the decision boundaries of TM, perceptron and a single layer neural network (SLNN) with ReLU activation. \\
Figure~\ref{fig:vis} shows the visualization of decision boundaries for TM, perceptron and SLNN obtained from testing on the Iris dataset. Each model was trained on the Iris dataset for $50$ epochs. The TM consisted of $50$ clauses and the SLNN was made up of $50$ neurons in the hidden layer activated by the ReLU function. The TM, perceptron and SLNN obtained $96\%$, $89\%$ and $94\%$ accuracy on the testing set, respectively\footnote{The code to reproduce these results and visualize decision boundaries is available here: github}. \\
As we can see from Figure \ref{fig:sfig1}, the decision boundaries of TM are cumulatively formed from clauses. Each clause contributes to the decision boundary as $C_i = \{0,1\}$. Basically, the decision boundary formed is in the form of \emph{steps} by filled contour lines. TM's decision boundary is like an unsmooth approximated version of a non-linear neural network, shown in Figure~\ref{fig:sfig3}. However, in case of a new data point, the entire curved decision boundary of a neural network might have to be changed in order to accommodate for the new data point. Whereas for TM, a clause (or a set of clauses) can simply learn the new pattern and create a \emph{step} in the decision boundary to include the new data point in the correct region. Note that decision boundaries can vary depending upon initialization and learning trajectory of TM.\\
To show that TM has the capability to separate non-linearly separable patterns, we show a toy like example, namely the TM's decision boundaries for the XOR problem in Figure \ref{fig:xor_boundary}. The perceptron or linear neural networks are incapable of solving the XOR problem. However, as can be seen from Figure~\ref{fig:xor_boundary}, TM can create decision boundaries that separate such patterns courtesy of the clauses. Each clauses contributes in the construction of the decision boundary. Here, we use $4$ clauses to learn the XOR sub-patterns. The theoretical analysis of convergence of TM on the XOR problem can be found in~\cite{jiao2021convergence}.

\subsection{Interpretability}
As previously explained, each propositional expression is a conjunctive clause, consisting of feature, 
in their original or negated forms, interacting with each other using logical \emph{AND} operations. 
These clauses can form a simplified representation of the arm selection policy  by combining them into 
a single Disjunctive Normal Form (DNF) expression. Since clauses are assigned to each class of the 
multiclass problem, we can produce a single DNF expression for each class. These DNF expressions are 
propositional logic expressions made up of binarized features. The TM is able to produce these 
interpretations demonstrating how it interprets the context with respect to each arm.\\
Here, we show the simplified propositional expressions for each class, obtained from TM trained on the Iris dataset:
\begin{enumerate}
\item
\begin{enumerate}
    \item[\emph{\textbf{Class-1:}}] $x_{10} \lor x_{14} \lor x_{15} \lor x_3$
    \item[\emph{\textbf{Class-2:}}]  $\neg x_1 \lor x_{12} \lor \neg x_{13} \lor \neg x_{14} \lor x_{16} \lor x_8 \lor \neg x_9 \lor (x_{10} \land x_{11} \land x_{15} \land x_2 \land \neg x_4 \land x_5 \land x_6 \land \neg x_7) \lor (x_{10} \land x_{11} \land x_{15} \land x_3 \land \neg x_4 \land x_5 \land \neg x_7)$ 
  \item[\emph{\textbf{Class-3:}}]  $~\neg x_{10} \lor \neg x_{11} \lor \neg x_{15} \lor (x_1 \land \neg x_{12} \land x_{13} \land x_{14} \land \neg x_{16} \land x_2 \land x_3 \land x_5 \land \neg x_8 \land x_9) \lor (\neg x_{12} \land \neg x_{16} \land x_4)$
\end{enumerate}
\end{enumerate}
The above expressions are obtained by combining the top ten highest weighted positive clauses for each class by \emph{OR}-ing them and just simplifying the Boolean expressions\footnote{Requires a couple of lines of code using the Sympy library.}.

\section{Discussions}
\subsection{Tsetlin Machine vs Deep Learning}
TM is a machine learning algorithm based on propositional Boolean expressions and logical operations. It is able to compete in performance with much larger deep learning models containing hundreds of thousands to many millions of floating point parameters. Different from deep neural networks, TM consists of a few thousands of binary clauses~\cite{dctm,rohan}, resulting in simplicity and low  memory and energy consumption~\cite{newcastle}. This feature makes TM suitable for mobile computing and federated learning in power constraint IoT devices. In addition, the propositional-logic based clauses are more interpretable than float-number based operations utilized in deep neural networks.\\   
Similar to deep learning, TM is also prone to overfitting. In other words, TM also learns patterns related to noise in the training data. For example,  TM is able to achieve $100\%$ training accuracy on large datasets like CIFAR-100, but the performance drops during validation . To mitigate the overfitting problem, a new version of TM, called the Drop Clause TM ~\cite{dctm}, has been proposed, which reduces redundancy and improves the generalization capabilities of TM.


\subsection{Tsetlin Machine vs Perceptron}
\begin{figure}[!h]
\centering
\begin{subfigure}{\columnwidth}
\centering
    \includegraphics[width=\columnwidth]{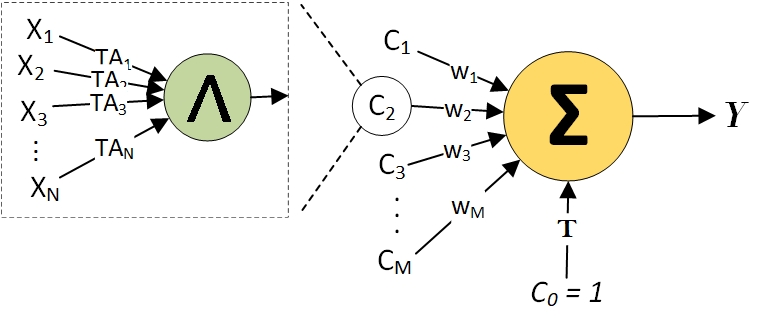}
    \caption{TM}
    \label{fig:tm}
\end{subfigure}
\begin{subfigure}{0.6\columnwidth}
\centering
    \includegraphics[width=\columnwidth]{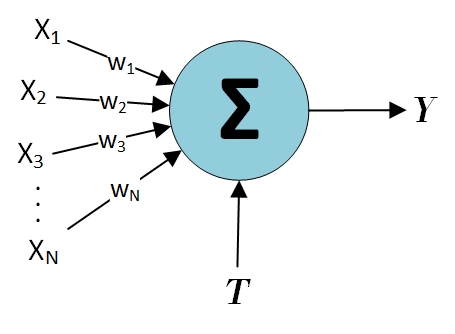}
    \caption{Perceptron}
    \label{fig:per}
\end{subfigure}
\caption{Tsetlin Machine vs Perceptron}
\label{fig:tmvsp}
\end{figure}

As shown in Section \ref{sec:w}, the weight update in TM can be considered as a special case of the perceptron learning algorithm for binarized input. Additionally, TM has a phase of learning in the clause level, i.e. representing a sub-pattern by a clause, which gives it better representability and flexibility. Figs. \ref{fig:tm} and \ref{fig:per} show the difference in structures of the TM and perceptron. The number of learnable parameters in the perceptron is restricted by the number of inputs. On the contrary, in TM, the number of clauses can be configured independently to the input's dimension, giving it structural flexibility.\\
Based on the descriptions in Section \ref{sec:TM}, we understand that the operational concept of the T is modularized, which has three parts. Firstly, distinct clauses learn various local sub-patterns. Secondly, the weights are assigned to these clauses according to their importance in solving the task. Finally,  these sub-patterns are combined for the final classification. \\ 
On the other hand, the perceptron has to learn global patterns directly from data in a single phase, limited to solving linearly separable patterns, whereas TM has demonstrated its competence in solving much more complex problems~\cite{dctm,rohan1,rohan3,rohan,granmo2019convtsetlin,abeyrathna2020intrusion,abeyrathna2020nonlinear,berge2019text}, which traditional machine learning algorithms are incapable of.

\section{Conclusions}
\label{sec:conclusion}
In this paper, we try to bridge the gap between the perceptron (and single-layer neural network) and TM by showing the similarities that lie in their respective structure and learning procedures. We formalize the learning mechanism of TM by dividing the learning phase into two layers. We show the equivalence of TM's weight update phase with the perceptron learning algorithm. An empirical analysis and visualization of decision boundaries demonstrates how TM can solve nonlinearly separable patterns, like the XOR problem, which the perceptron (and linearly activated SLNN) is incapable of. The decision boundaries show similarities to that of the single-layer neural network. Apart from visualization of decision boundaries, the empirical analysis also shows the flexibility of determining clause length, memory consumption, convergence rate and obtaining interpretable Boolean rules.

\bibliographystyle{named}
\bibliography{ijcai22}


\appendix

\section*{Appendix}

\section{Convergence of Perceptron}
\label{sec:app}
Let $W=[w_1,...,w_n]^T$ be the weights of the perceptron and $X^{n\times D}=[(x_1,y_1),...,(x_n,y_n)]^T$ be the input-label pairs, where $y_i \in \{-1,1\}$ and each $x_i$ is a $D$-dimensional vector. Assumptions:
\begin{enumerate}
    \item There exists some $W^*$ such that $\Vert W^* \Vert = 1$, and for some $\gamma > 0$, $\forall n = 1 \dots N$:
        \begin{equation}
            y_{n}^{} x_{n}^T W^* \geq \gamma
        \end{equation}
    \item Also assume, $\forall n$:
        \begin{equation}
            \Vert x_n \Vert \leq R
        \end{equation}
\end{enumerate}
\paragraph{\emph{Perceptron Convergence:}}The number of updates $k$ required for the perceptron to converge to a local minimum is bounded by:
\begin{equation}
    k \leq \frac{R^2}{\gamma^2}
\end{equation}
\paragraph{\emph{Proof:}}Let $W_k$ be the weight vector after $k^{th}$ update and $\Vert W_1 \Vert = 0$. So, for $k+1$ we have:
\begin{align*}
    W_{k+1} & = W_k + y_nx_n \\
    W_{k+1}^TW^* & = (W_k + y_nx_n)^TW^* \\
     & = W_k^TW^* + y_n^{}x_n^TW^*\\
     & \geq W_k^TW^* + \gamma
\end{align*}
It follows by induction on $k$ that:
\begin{equation*}
    W_{k+1}^TW^*\geq W_k^TW^* + \gamma \geq W_{k-1}^TW^* + 2\gamma \geq \dots \geq k\gamma
\end{equation*}
In addition, as $\Vert W_{k+1} \Vert \Vert W^* \Vert \geq W_{k+1}^TW^*$, then we have:
\begin{equation}
\label{eq:ww}
    \Vert W_{k+1} \Vert \geq k\gamma
\end{equation}
Now, we can also write:
\begin{align*}
    \Vert W_{k+1} \Vert ^2 & = \Vert W_k + y_nx_n \Vert ^2 \\
    & = \Vert W_k \Vert ^2 + y_n\Vert x_n \Vert ^2 + 2y_n^{}x_n^TW_k \\
    & \leq \Vert W_k \Vert ^2 + R^2
\end{align*}
It follows by induction on $k$ that:
\begin{equation}
\label{eq:ww2}
    \Vert W_{k+1} \Vert ^2 \leq kR^2
\end{equation}
Combining equations \ref{eq:ww} and \ref{eq:ww2}, we have:
\begin{align}
    k^2\gamma^2 \leq \Vert W_{k+1} \Vert ^2 \leq kR^2 \\
    \label{eq:final}
    k \leq \frac{R^2}{\gamma^2}
\end{align}
This shows that the number of update steps, $k$, required by the perceptron to obtain a local minimum, is bounded. In case of TM, the same proof above holds with only one difference: $\forall n$, $x_n \in \{0,1\}$, i.e. the input is a binary vector. As the (Frobenius) norm of a binary vector is the square root of the number of non-zero elements and the number of non-zero elements can be atmost $D$. So, the second assumption becomes:
\begin{equation}
    \Vert x_n \Vert \leq \sqrt{D}
\end{equation}
So, for TM, Equation~(\ref{eq:final}) becomes:
\begin{equation}
    k \leq \frac{D}{\gamma^2}
\end{equation}

\end{document}